\documentclass{article}

\usepackage{arxiv}

\usepackage[utf8]{inputenc} % allow utf-8 input
\usepackage[T1]{fontenc}    % use 8-bit T1 fonts
\usepackage{hyperref}       % hyperlinks
\usepackage{url}            % simple URL typesetting
\usepackage{booktabs}       % professional-quality tables
\usepackage{amsfonts}       % blackboard math symbols
\usepackage{nicefrac}       % compact symbols for 1/2, etc.
\usepackage{microtype}      % microtypography
\usepackage{lipsum}
\usepackage{graphicx}

\newcommand{\code}[1]{\texttt{\detokenize{#1}}}

\title{Eisen: a python package for solid deep learning}

\author{
  Frank Mancolo \\
  The Eisen Project \\
  Cambridge, MA 02139
}

\begin{document}
\maketitle

\begin{abstract}
Eisen is an open source python package making the implementation of deep learning methods easy. It is specifically tailored to medical image analysis and computer vision tasks, but its flexibility allows extension to any application. Eisen is based on PyTorch and it follows the same architecture of other packages belonging to the PyTorch ecosystem. This simplifies its use and allows it to be compatible with modules provided by other packages. Eisen implements multiple dataset loading methods, I/O for various data formats, data manipulation and transformation, full implementation of training, validation and test loops, implementation of losses and network architectures, automatic export of training artifacts, summaries and logs, visual experiment building, command line interface and more. Furthermore, it is open to user contributions by the community. Documentation, examples and code can be downloaded from \url{http://eisen.ai}.
\end{abstract}


\section{Introduction}

Deep learning methods have recently emerged as extremely effective tools to solve challenging tasks in a wide range of domains. Computer vision has been revolutionized by the introduction of deep learning approaches. Medical image analysis, as a branch of computer vision, has not been an exception: the vast majority of recently published papers in this domain is based on deep learning and the use of convolutional neural networks (CNNs).

Most open source implementations of recent methods are built from scratch. Almost every published work leverages custom made data processing routines as well as custom training, validation and testing loops and often relies on personalized model implementations. This makes development of new approaches a labor intensive procedure. Furthermore, whenever an existing approach is applied to a new dataset or whenever it is necessary to change part of an existing implementation to accommodate new functionality, multiple changes need to be made to the code base and, depending on its architecture, silent bugs and unnecessary complexity might appear.

Performance benchmarks, which are crucial to reveal the true capabilities of newly proposed approaches, are also not easily obtainable and often misleading, as researchers have often to rely on their own implementations of, at least, part of the comparing method implementation. This is often error prone since this code is often untested and some implementation detail can be missed by even expert programmers. 

Finally, whenever research changes hands and project owners with different background and skills follow one another, these custom architectures need to be re-evaluated, understood and hopefully expanded in order to incorporate new work. Convoluted custom designs represent a big disadvantage in this scenario as creating new compatible code and reusing existing modules is challenging and often impossible.

These are just some of the issues motivating the development of Eisen. Our goal is to provide an "opinionated" framework for the development of deep learning approaches for vision tasks. In this sense, Eisen proposes a modular architecture where each module performs a very limited amount of functionality through a standard interface and a very readable implementation. Modules can be mixed and matched, each with its specific role, and used during training, validation, testing as well as model serving in order to realize complex functionality. Each module is easy to read, is documented and follows a standard implementation. This results in an architecture which is straightforward to understand, fits the design principles behind PyTorch and its ecosystem, and offers flexibility and opportunities for extension. 

Eisen modules implement various aspects of deep learning model development and experimentation: dataset reading capabilities in order to bring data from public challenges or other sources into Eisen; data loading in various format such as Nifti, ITK, PNG, JPEG, DICOM and others; data manipulation and augmentation, in order to convert data format, crop, adjust, re-sample images; automatic export of training artifacts, Tensorboard summaries, detailed logs of workflow activity; several network architectures such as UNet \cite{ronneberger2015u}, VNet \cite{milletari2016v}, Highres3DNet \cite{li2017compactness}, Obelisk \cite{heinrich2018obelisk}; additional layers, losses and metrics; and a number of other minor features.

Eisen's functionality can be accessed by either importing the relative modules in any python project via a simple \code{import eisen}, or through the CLI, or even directly by copying from the source code from our GitHub repository. Eisen core functionality is offered under MIT license. Other functionality that is not part of the core is offered under GNU GPL v3.0 license. 

\section{Related work}

Various recent projects have attempted to motivate researchers to develop their methods using shared implementations and common code bases. These approaches have often lacked flexibility and were not generic enough to accommodate the ever-changing landscape of datasets and methods emerging from the community. A few solutions were over-engineered, and required users to make several changes to the code in order to be able to use their own models or data loading routines; others were incomplete and focusing only on some aspect of the problem such as data manipulation or data iteration loops; a few failed to gain traction among the community.

NiftyNet \cite{gibson2018niftynet} is one of the most notable and popular software packages for easy implementation of deep learning approaches in medical field. It has been initially published in 2018, and is based on Tensorflow 1.x. Its architecture follows well known software engineering patterns and it includes multiple popular models such as 3D-Unet \cite{cciccek20163d}, V-Net \cite{milletari2016v}, etc.
DLTK \cite{pawlowski2017dltk} represents an alternative package, which has similar functionality. It was initially published in 2017, and is also based on Tensorflow 1.x. It as well includes data pipelines and training capabilities with different state of the art models which are tailored for medical image analysis. Other efforts are represented by the "medicaltorch" (\url{https://github.com/perone/medicaltorch}) package, which is based on PyTorch and includes crucial functionality to deal with data loading, augmentation, pre-processing and training itself. A similar package is "batchgenerators" (\url{https://github.com/MIC-DKFZ/batchgenerators}) which extends standard PyTorch classes to provide ability to load and manipulate data. TorchIO \cite{perez2020torchio} implements very useful transformations to perform data augmentation and manipulation with the capability of efficiently dealing with volumetric data types. Additionally it allows patch-wise volume processing in order to overcome the memory constraints of current hardware accelerators for DL. Finally, \cite{mehrtash2017deepinfer} and \cite{milletari2018tomaat} propose approaches to allow medical image analysis model serving over the network through flexible interfaces. Their goal is to make trained models accessible for inference through common visualization tools.

\section{Architecture}
Eisen is currently implemented by two different Python packages, \code{eisen_core} and \code{eisen_cli}, that can be installed on their own or together through a meta-package containing both. A third Python package is currently planned in our roadmap in order to bring model serving capabilities to the Eisen ecosystem. The most complete installation of Eisen can be obtained by executing \code{pip install eisen} which installs the aforementioned meta-package. 

Comprehensive documentation about Eisen can be obtained at \url{http://docs.eisen.ai}.

\begin{figure*}[!t]
\begin{center}
   \includegraphics[width=250px]{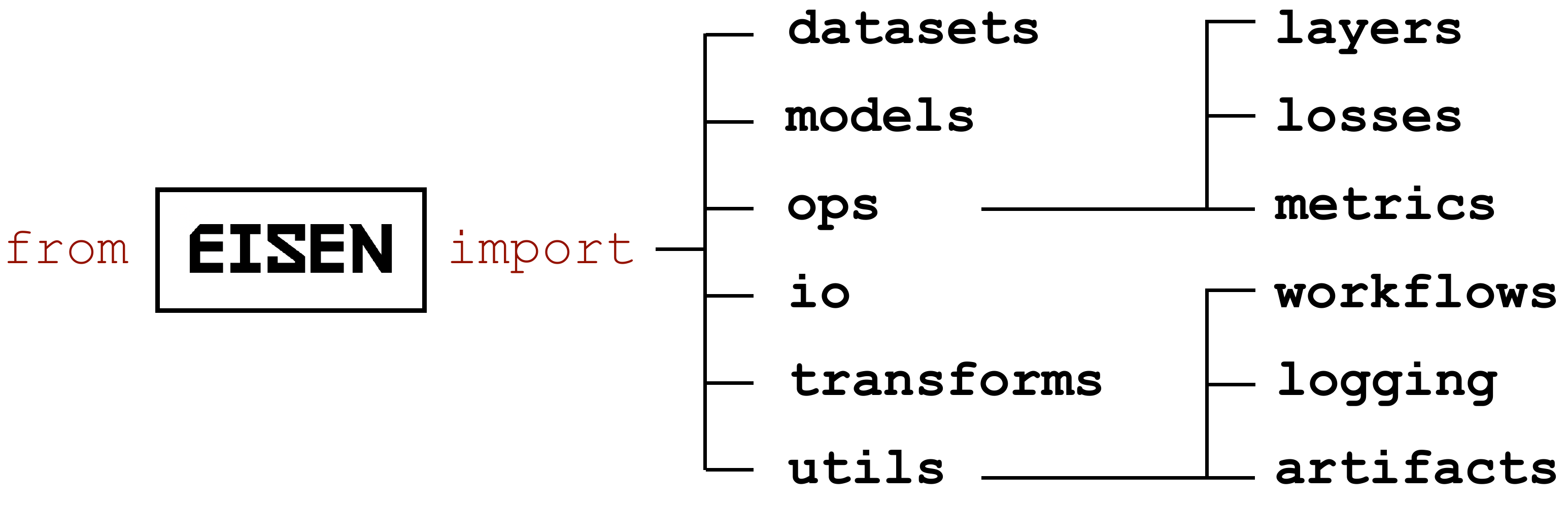}
\end{center}
\caption{Organization of our Python package. The package structure is inspired by the modules in torchvision.}
    \label{fig:organization}
\end{figure*}

\subsection{Core functionality}

The core functionality of Eisen can be imported through a simple 
\code{import eisen}. The package organization is inspired by torchvision, which is one of the most widespread and appreciated packages for computer vision in the PyTorch ecosystem. Additionally, most modules that are part of torchvision are also compatible with Eisen. The same is true for most modules implemented in other packages that are compatible with PyTorch.  

A graphical representation of the module organization of Eisen is shown in Figure \ref{fig:organization}. Details about the content and role of each module are discussed below. Since it would be impractical to depict every single module in our library in a single picture, especially considering that our package is under active development, Figure \ref{fig:organization} omits individual models. Documentation about each module can be found at \url{http://docs.eisen.ai/eisen/api.html}.

\subsubsection{Datasets}
Deep learning method are able to solve tasks by learning model parameters that optimize a loss function according to the content of a dataset. Data is therefore central to any deep learning model.

Eisen includes functionality to handle different standard popular and often public datasets that can be obtained from the internet,
it also includes the capability of supplying own datasets which can contain arbitrary data. 

Personalized datasets can be indeed created by specifying their content in a JSON file. The expected JSON file structure is a list of structures. Each entry of the list contains one element of the dataset. Each field of the structure stores different information about that data point. For example, a dataset containing images and labels would be represented in JSON as a list of structures having fields 'image' and 'label' each storing respectively one image and relative label. 

Every time an item that belongs to the dataset is requested, a Python dictionary is returned. The dictionary has multiple fields each storing information relative to the returned data-point. It is important to note that data is always handled via dictionaries in Eisen. This concept is useful when considering the way transformations and I/O objects are implemented.

The datasets currently supported by Eisen Core (v0.0.4) are summarized in Table \ref{tab:dataset}. Datasets implemented in other python modules such as torchvision can be adapted to Eisen using the \code{EisenDatasetWrapper} which is available in the \code{utils} submodule.

\subsubsection{I/O}
Eisen I/O functionality is contained in the eisen.io module. I/O functionality is implemented by transforms. That is, this
functionality behaves just like any other eisen.transform module. The only difference is that I/O operations involve disk.
Another reason of this distinction is that we decided to follow the package structure of torchvision, which has an
torchvision.io sub-package.

I/O modules are designed to receive an element of the dataset and update it with data objects resulting from loading files from disk.
This is often the first step of complex pipelines and, since all transforms and I/O modules affect the data dictionary in a similar way, whenever the task at hand change resulting in different data types being loaded, it is sufficient to use a different reader to get the data into Eisen in just the same way as before. 

\begin{table}[t!]
\caption{Eisen Core v0.0.4 supports both generic (via JSON file) and publicly available datasets that have been used in challenges and for benchmarks of deep learning models.}
\centering
\begin{tabular}{@{}ll@{}}
\toprule
Dataset Name                                       & Module (\code{eisen.datasets.} prefix)   \\ \midrule
Medical segmentation decathlon \cite{simpson2019large}                    & \code{MSDDataset}    \\
Patch Camelyon Dataset \cite{veeling2018rotation}                           & \code{PatchCamelyon} \\
CAMUS Ultrasound Dataset \cite{leclerc2019deep}                          & \code{CAMUS}          \\
Intracranial Hemorrhage Detection Challenge (RSNA) & \code{RSNAIntracranialHemorrhageDetection} \\
Bone Age Estimation Challenge (RSNA)               & \code{RSNABoneAgeChallenge} \\
Personalised dataset as JSON file                  & \code{JsonDataset}  \\ \bottomrule
\end{tabular}
\label{tab:dataset}
\end{table}

\subsubsection{Transforms}

Transforms, similarly to I/O modules, operate as well on data dictionaries and implement functionality such as pre-processing and augmentation of data.
Transforms are operations that can be composed together by using \code{torchvision.transforms.Compose}. Once the transforms are composed together they realize a transformation chain.

Each transform should implement one, and just one, basic operation.
Since arguments are not explicitly copied by transforms
there is no dramatic overhead brought by the presence of multiple transforms which implement a simple operation, and are stacked together to realize complex functionality. 

This paradigm is very different from what can be observed in many open source method implementations, which propose complex data handling pipelines implementing the whole in a single, rather long, function. We believe that this coding practice is bug prone, inelegant, results in non-modular code and prevents module reuse, therefore we strongly discourage it within Eisen.

Users are on the other hand encouraged to implement their own transforms. These can be, of course, imported within the code and can also be made accessible via Eisen configuration files and accessed via
CLI as soon as they are added to the \code{PYTHON_PATH}. User implementations need to stick to the basic transform template, which is not formally defined in code but is explained in the documentation
at \url{http://docs.eisen.ai/eisen/api.html#module-eisen.transforms}.

Transformations implemented in other python modules such as torchvision and TorchIO can be adapted to Eisen using the \code{EisenTransformWrapper} which is available in the \code{utils} submodule.

\subsubsection{Models}
In order to favor reproducibility of deep learning approaches and easy benchmarking, as well as providing "starter-kit" tools to users approaching a certain problem for the first time, we include several well-known neural network architecture within Eisen. This is similar to the approach taken by torchvision which ships network architectures for classification, segmentation and beyond within the package. 

Eisen Core v0.0.4 includes various implementations that are summarized in Table
\ref{tab:networks}. Models implemented in other python modules such as torchvision can be adapted to Eisen using the \code{EisenModuleWrapper} which is available in the \code{utils} submodule.

\begin{table}[]
\caption{Eisen Core v0.0.4 implements several well known models for medical image analysis. Code is often obtained by other open source repositories under MIT License.}
\centering
\begin{tabular}{@{}ll@{}}
\toprule
Network architecture & Module  (prefix \code{eisen.models.}) \\ \midrule
3D-UNet \cite{cciccek20163d}             & \code{UNet3D}                        \\
UNet \cite{ronneberger2015u}                & \code{UNet}                          \\
VNet \cite{milletari2016v}                & \code{VNet}                          \\
Obelisk \cite{heinrich2018obelisk}             & \code{ObeliskMIDL}                   \\
HighRes3DNet \cite{li2017compactness}        & \code{HighRes3DNet}                  \\
HighRes2DNet \cite{li2017compactness}        & \code{HighRes2DNet}                  \\ \bottomrule
\end{tabular}
\label{tab:networks}
\end{table}

\subsubsection{Ops}
We include various operations that are useful when developing deep learning models. The operations are always implemented in PyTorch and derived from the class
\code{torch.nn.Module} as suggested by the PyTorch documentation itself. Eisen contains implementations of layers, metrics and losses. Losses and metrics implementation include methods such as the Dice loss \cite{milletari2016v}, which find useful application especially in tasks belonging to the medical domain.

\subsubsection{Utils}
The \code{eisen.utils} sub-module contains a few useful objects such as wrappers that have the ability to extend the compatibility of third party modules to Eisen. Moreover, it contains workflows, artifact generation utilities and hooks having the ability to monitor the workflows and save summaries as well as artifacts as a result.

\paragraph{Workflows}
Workflows implement the core functionality of training, testing and validation. These modules are documented at the URL
\url{http://docs.eisen.ai/eisen/api.html#workflows} and contain complete yet generic implementations that
can be used to train, test and validate models tackling virtually any task: 2D segmentation, 3D segmentation,
classification, regression, pose estimation and other tasks can all be handled by standard Eisen workflows.

The content of data batches is automatically routed to the correct \code{torch.nn.Module} for processing. The result
of such processing can be automatically interpreted and logged in the tensorboard, the console, logfiles, etc. by using hooks. 

Eisen workflows resolve the pain of having to write training, validation and testing loops from scratch each time. At the same time workflows impose a standard and a structure that brings together all the pieces composing a development and experimentation pipeline. This
does not mean that users are barred from proposing their own workflows. For example, Eisen Core v0.0.4 does not support
training reinforcement learning agents due to the absence of a workflow that can cope with reinforcement learning
environments. Users can in this case implement their own workflow according to what stated in the documentation at
\url{http://docs.eisen.ai/eisen/api.html#workflows} and what can be learned from Eisen source code at \url{http://github.com/eisen-ai}.

\begin{figure*}
\begin{center}
   \includegraphics[width=\textwidth]{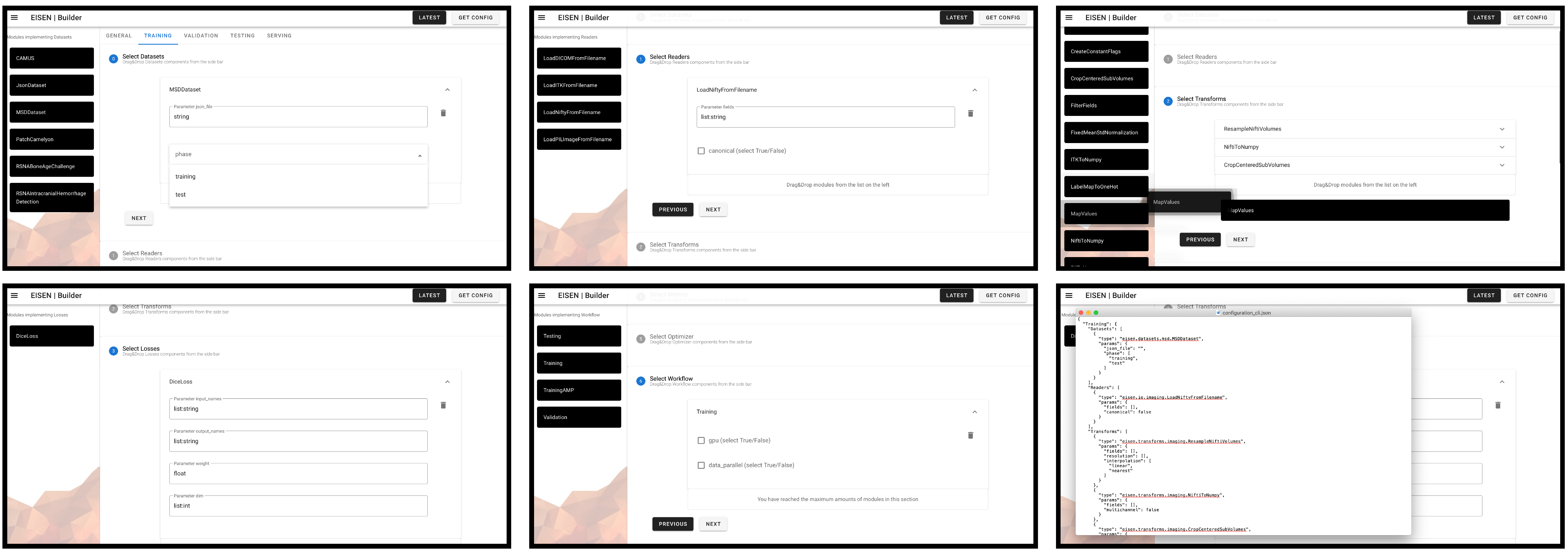}
\end{center}
\caption{Example of Eisen Builder user interface (\url{http://builder.eisen.ai}) which helps users obtain configuration files to execute workflows via CLI. Modules can be dragged and dropped into work area to realize the desired functionality. This user interface can also be used as an excellent tool to investigate the API of implemented modules. This image is best viewed digitally.}
    \label{fig:ui}
\end{figure*}

\paragraph{Logging}
The activity of a workflow can be logged by using "Hooks". Hooks in Eisen listen for certain events to be generated
by a specific workflow. They respond to these events by performing actions on the data provided as a result of said
events.

For example, as the training workflow iterates the dataset to obtain batches of data to optimize the neural network,
the result of the optimization, its inputs, outputs, losses, metrics, a model snapshot and the current epoch number
are collected and recorded. Once an epoch finishes, a signal is generated. The collected information is therefore
sent to the hooks which listen to the specific workflow.

Each hook can perform its own functionality on the resulting data. For example \code{eisen.utils.logging.LoggingHook} logs
the progress of training in terms of losses and metrics on the console. \code{eisen.utils.logging.TensorboardSummaryHook}
exports inputs, outputs, losses and metrics to the tensorboard in a completely automated way. Importantly, tensorboard summaries are saved automatically by automatically inferring data types and in order to offer as many insights about the monitored workflow as possible. All of this, without any need to write code. 

Some hooks are also able to save artifacts based on whether the best metric or loss has been reached by the model.
These hooks such as \code{eisen.utils.artifacts.SaveTorchModelHook} or \code{eisen.utils.artifacts.SaveONNXModelHook} save
a snapshot or even the whole model history (the sequence of best models over time) in PyTorch and ONNX format respectively.

Users can also define their own hooks by referring to the source and the documentation as implementation guidance.

\paragraph{Artifacts}
Artifacts can also be saved without using hooks (see above). It is possible to use eisen modules to serialize
models to disk. In Eisen v0.0.3 it is possible to save models to disk in PyTorch or ONNX format. This functionality
is realized by \code{eisen.utils.artifacts.SaveTorchModel} and \code{eisen.utils.artifacts.SaveONNXModel}.

These modules can be used when the user wants more control on how models are saved. They are also useful when
models are trained without using Eisen workflows. In that case, in fact, it will not be possible to use hooks during
training, validation or testing. I will therefore be impossible to use modules such as
\code{eisen.utils.artifacts.SaveTorchModelHook} or \code{eisen.utils.artifacts.SaveONNXModelHook}.

Transforms are instantiated to manipulate the data as represented in the dataset (instance of \code{torch.utils.data.Dataset}) instance. For example they load from disk, re-sample, crop, pad and normalize the images. A model is also instantiated (instance of \code{torch.nn.Module}). A list of losses and metrics is supplied (instance of \code{list} of \code{torch.nn.Module}). Finally the optimizer is instantiated in order to optimize the model parameters (instance of \code{torch.optim.Optimizer}); note that the optimizer might not be necessary depending on the workflow. A workflow using all these module instances of is finally instantiated and run. While running it emits events that are in turn picked up by "hook" instances that perform actions such as logging, artifact saving and summary export.

\subsection{Command line interface}

Eisen can be accessed via a command line interface (CLI). Eisen-CLI can be installed by either using the meta package via \code{pip install eisen} or explicitly via \code{pip install eisen_cli}.

Users can train, test and validate models via CLI. In order to use the CLI, it is necessary to provide a configuration file for Eisen in JSON format. This file contains instructions about what to do during training, validation and testing. It contains a description of what modules should be instantiated and what transformations and datasets should be used during each phase. Any module, including third party modules or user modules can be used in the configuration. 

The simplest way to obtain the configuration is to use http://builder.eisen.ai which provides a visual interface to build complex configuration files leveraging Eisen modules and functionality. An example of this is shown in Figure \ref{fig:ui}. This user interface (UI) does not currently include user-defined transforms and third party modules, which still need to be brought into the configuration manually if desired. Once users are acquainted with the way Eisen CLI works and how configuration files are structures, it will be easy for them to manually edit the resulting JSON files.

\section{Using Eisen}
Having a clear functionality of each block within Eisen, we can run experiments and development of new approaches by joining together (either via python code or JSON configuration file) the various module instances within a workflow that can be run through a single line of code or via a single command via the CLI.

\begin{figure*}[!t]
\begin{center}
   \includegraphics[width=\textwidth]{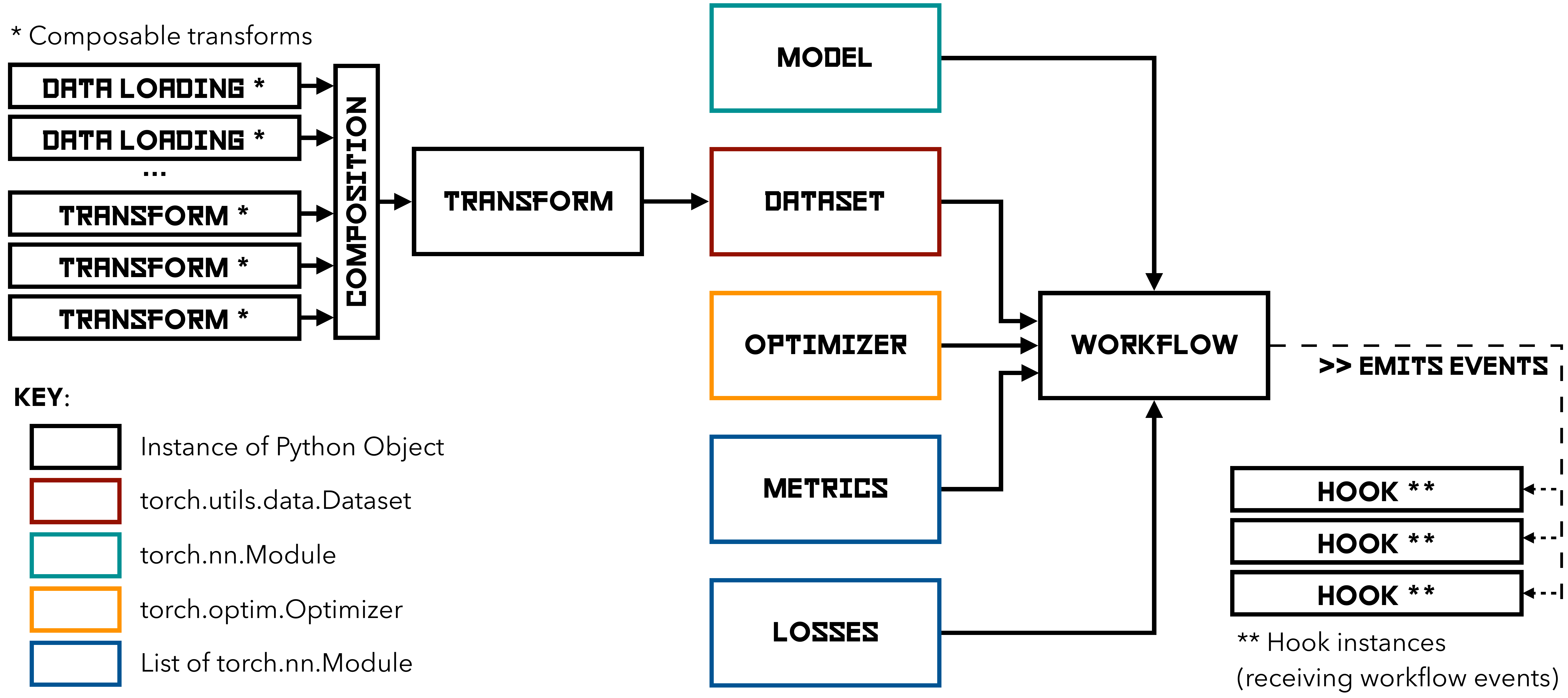}
\end{center}
\caption{A schematic representation of Eisen's operational architecture. Color convention indicates instance types.}
    \label{fig:architecture}
\end{figure*}

A schematic representation of how Eisen works is shown in Figure \ref{fig:architecture}. 

We refer the reader to additional resources and tutorials that can be reached at \url{http://docs.eisen.ai/eisen/tutorials.html} and are also summarized in Table \ref{tab:tutorials}. The example and tutorials can be directly run on GPUs using the Google Colab platform or downloaded as python notebook. They represent an excellent starting point to work with Eisen.

\begin{table}[]
\caption{Various tutorials leveraging Eisen have been proposed. They can all be run on GPUs using the Google Colab platform. This table is best viewed digitally.}
\centering
\begin{tabular}{@{}ll@{}}
\toprule
Tutorial title & URL (Colab) \\ \midrule
Minimal Example with Medical Segmentation Decathlon \\ A simplified example of Eisen on a real-life use case              & \url{http://bit.ly/2HjLlfh}                       \\
Volumetric Segmentation Medical Segmentation Decathlon \\ An example of Eisen on a real-life use case                 & \url{http://bit.ly/39veXlZ}                          \\
MNIST classification using torchvision components with Eisen \\ Showcasing the capability of Eisen to work with other packages                 & \url{http://bit.ly/37oBdMZ}                          \\
MNIST classification using torchvision and Eisen with automatic \\ mixed precision (AMP) via NVIDIA Apex for PyTorch             & \url{http://bit.ly/2Q69Cdy}                   \\
Demonstration of the Eisen TensorboardSummaryHook exporting \\ workflow data (inputs, outputs, metrics and losses) to tensorboard        & \url{http://bit.ly/2IERhjA}                  \\
\bottomrule
\end{tabular}
\label{tab:tutorials}
\end{table}

\section{Conclusions}
Eisen is a Python packaged that implements an opinionated yet flexible framework for easy development and experimentation of deep learning models for computer vision. In particular, Eisen focuses on medical image analysis as it provides modules to read and manipulate volumetric data as well as data formats such as DICOM, Nifti and ITK that are typical of the medical domain. Eisen's purpose is to ease development, help developers at any level of knowledge and skill to work with deep learning, avoid duplicate efforts, offer an iron-solid foundation for small and big projects involving imaging. We believe Eisen offers unprecedented flexibility and user friendliness. We would like to encourage the community to contribute and get in touch with us on Slack. All information can be found on the official website \url{http://eisen.ai}.

\bibliographystyle{unsrt}  
\bibliography{references}  

\end{document}